\pdfoutput=1

\documentclass[11pt]{article}

\usepackage{ACL2023}

\usepackage{times}
\usepackage{latexsym}
\usepackage{booktabs}
\usepackage{multirow}

\usepackage{array}
\usepackage{caption}
\usepackage{tabularx}
\usepackage{colortbl}
\usepackage{longtable} 
\usepackage[lofdepth,lotdepth]{subfig}
\usepackage{graphicx}
\usepackage{xcolor} 
\usepackage{float}
\usepackage{makecell}
\usepackage{url}
\usepackage{hyperref}
\usepackage{amssymb}
\usepackage{amsmath}

\usepackage[T1]{fontenc}

\usepackage[utf8]{inputenc}

\usepackage{microtype}

\newcolumntype{C}[1]{>{\centering\arraybackslash}m{#1}}
\newcolumntype{L}[1]{>{\arraybackslash}m{#1}}
\newcommand{\textarrow}{$\hookrightarrow$}

\title{Sources of Noise in Dialogue and How to Deal with Them}

\author{Derek Chen \\
Columbia University, NY \\
\texttt{dc3761@columbia.edu} \\
\And
Zhou Yu \\
Columbia University, NY \\
\texttt{zy2461@columbia.edu} \\}

\begin{document}
\maketitle
\begin{abstract}
Training dialogue systems often entails dealing with noisy training examples and unexpected user inputs.  Despite their prevalence, there currently lacks an accurate survey of dialogue noise,  nor is there a clear sense of the impact of each noise type on task performance. This paper addresses this gap by first constructing a taxonomy of noise encountered by dialogue systems. 
In addition, we run a series of experiments to show how different models behave when subjected to varying levels of noise and types of noise.  Our results reveal that models are quite robust to label errors commonly tackled by existing denoising algorithms, but that 
performance suffers from dialogue-specific noise.  Driven by these observations, we design a data cleaning algorithm specialized for conversational settings and apply it as a proof-of-concept for targeted dialogue denoising. 
\end{abstract} 

\section{Introduction}
Quality labeled data is a necessity for properly training deep neural networks. More data often leads to better performance, and dialogue tasks are no exception~\citep{qian19metalearn}.  However, in the quest for more data, practitioners increasingly rely on crowdsourcing or forms of weak supervision to meet scaling requirements. 
Even when acting in good faith, crowdworkers are not trained experts which understandably leads to  mistakes.  This ultimately results in noisy \textit{training inputs} for our conversational agents. 
Moreover, when dialogue systems are deployed into the real world, they must also deal with noisy \textit{user inputs}.  For example, a user might make an ambiguous request or mention an unknown entity.  All these sources of noise eventually take their toll on model performance.

Before building noise-robust dialogue systems or denoising dialogue datasets, it would be helpful to know what types of noise exist in the first place.  Then our efforts can be spent more wisely tackling the sources of noise that actually make a difference.  Prior works have looked into
counteracting noisy user interactions~\citep{peng20raddle, liu20robustness}, but did not study the impact of noisy training data.  Moreover, they lack analysis on how noise influences performance across different model types or conversational styles.
Other works claim that dialogue agents can be easily biased by offensive language found in noisy training data~\citep{ung22saferdialogues, dinan20queens}. 
Given such a danger, we wonder ``How much toxic data actually exists in annotated dialogue data?''


\begin{figure}
  \includegraphics[width=\linewidth]{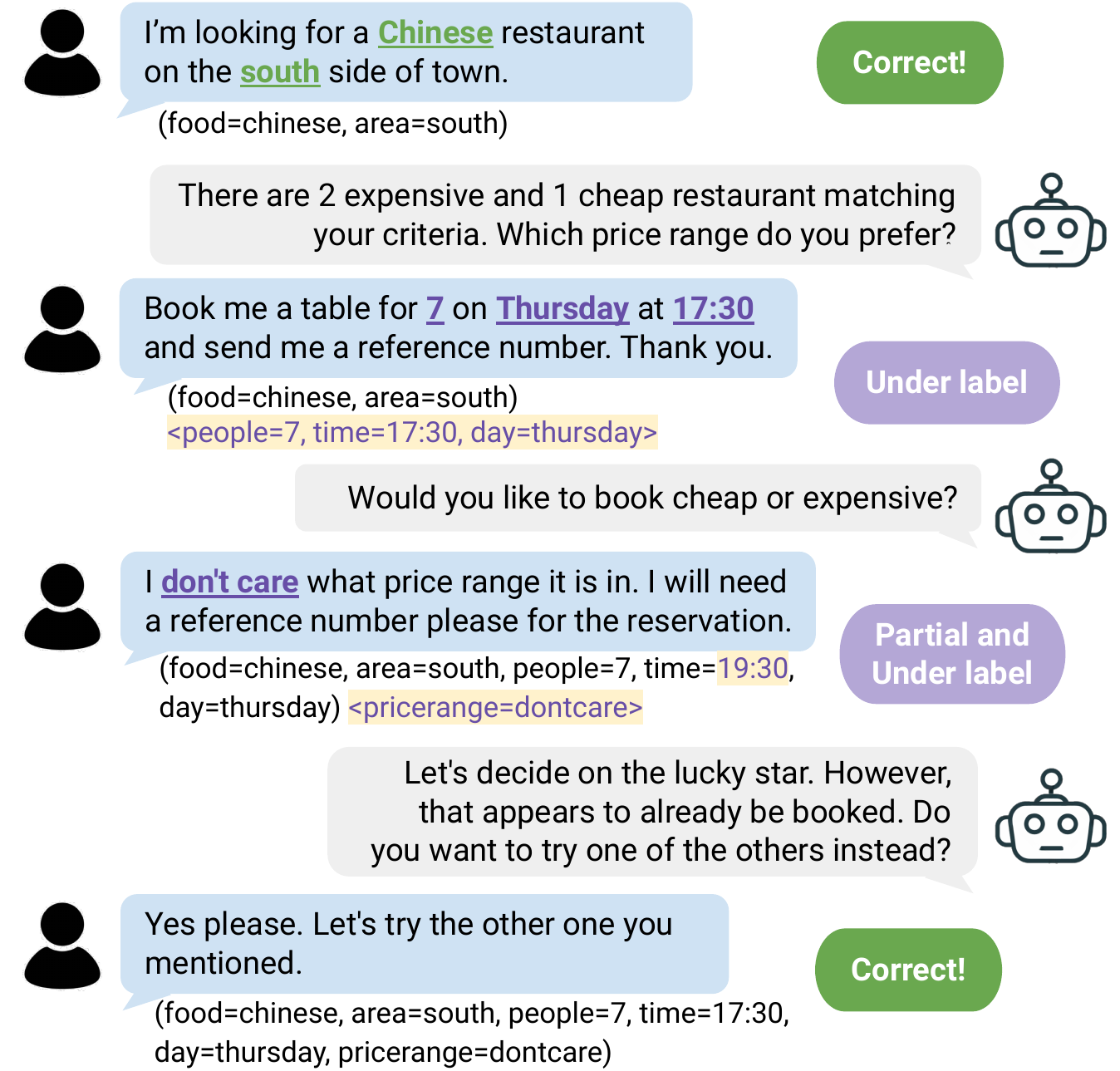}
  \caption{An example of label errors within MultiWoz 2.0 which contains partially filled and missing labels. We categorize this as two types of instance-level noise.}
  \label{fig:front_page}
\end{figure}

\begin{table*}[ht]
\centering
\resizebox{\textwidth}{!}{
{\begin{tabular}[width=\linewidth]{|l|c|cc|cccc|}
\hline & & & & & & &  \\[-2ex]
                &                & \textbf{Num.}     & \textbf{Collection} 
				& \textbf{Open}  & \textbf{Goal}     & \textbf{Synchronous} & \textbf{KB/} \\
\textbf{Dataset}& \textbf{Abbr.} & \textbf{Dialogs}  & \textbf{Methodology} & \textbf{Domain} 
                & \textbf{Oriented} & \textbf{Chat}         & \textbf{Document} \\
\hline
\small{Action-Based Conversations Dataset}
								& ABCD    &  10,042  & Expert Live Chat 	 &   & X & X & X \\
\rowcolor{lightgray}
DailyDialog 					& DD      &  13,118  & Post-conv Annotation  & X &   &   &   \\
Empathetic Dialogues      		& ED      &  24,850  & Live Chat         	 & X &   & X &   \\
\rowcolor{lightgray}
Google Simulated Conversations  & GSIM	  &  3,008   & Machine to Machine 	 &	 & X &   &   \\
Key-Value Retrieval for In-Car 	& KVRET   &  3,031   & Wizard of Oz 		 &   & X &   & X \\
\rowcolor{lightgray}
\small{Machine Interaction Dialog Act Schema}
								& MIDAS   &     468  & Live Chat 			 & X &   & X &   \\
MultiWoz 2.3 					& MWOZ    &  10,419  & Wizard of Oz 		 &   & X &   &   \\
\rowcolor{lightgray}
Schema Guided Dialogue 			& SGD     &  42,706  & Post-conv Annotation  &   & X &   &   \\
TicketTalk (TaskMaster 3) 		& TT      &  23,789  & Dialogue Self-Play 	 &   & X &   &   \\
\rowcolor{lightgray}
Wizard of Wikipedia				& WOW     & 22,311   & Wizard of Oz 	     & X &   & X & X \\
\hline
\end{tabular}}}
\caption{Breakdown of ten dialogue datasets used in constructing the noise taxonomy.  The datasets were chosen to span a wide variety of annotation schemes, task specifications and conversation lengths. KB/Document refers to a dataset containing an external knowledge base or document to ground the conversation.  (See Appendix~\ref{sec:datasets})} \label{tab:dst}
\end{table*}

To investigate these concerns, we survey a wide range of popular dialogue datasets and outline the different types of naturally occurring noise. 
Building on this exercise, we also study the patterns of annotation errors to determine the prevalence of each noise type and identify the most likely causes of noise.  Next, we run transformer models through the gamut to find out how well they handle the different types of 
noise documented in the previous step.  In total, we test 3 model types on 7 categories of noise across 10 diverse datasets spanning 5 dialogue tasks.  We discover that most models are quite robust to the label errors commonly targeted by denoising algorithms~\citep{natarajan13learning, reed14bootstrap}, but perform poorly when subjected to dialogue-specific noise.
Finally, to verify we have indeed identified meaningful noise types, we apply our findings to denoise a dataset containing real dialogue noise.  As a result, we are able to raise joint goal accuracy on MultiWOZ 2.0 by 42.5\% in relative improvement.


In total, our contributions are as follows:
(a) Construct a realistic taxonomy of dialogue noise to guide future data collection efforts. (b) Measure the impact of noise on multiple tasks and neural models to aid the development of denoising algorithms. (c) Establish a strong baseline for dealing with noise by resolving dialogue specific concerns, and verify its effectiveness in practice.


\section{Dialogue Datasets}
A data-driven taxonomy of dialogue noise was designed by manually reviewing thousands of conversations across ten diverse datasets and their accompanying annotations. 
The datasets were chosen from non-overlapping domains to exhaustively represent all commonly considered dialogue tasks.  At a high level, they are divided into six task-oriented dialogue datasets and four open domain chit-chat datasets.  The task-oriented datasets are comprised of MultiWoz 2.0 (MWOZ) \citep{budzianowski18multiwoz}, TicketTalk (TT) \citep{byrne19taskmaster}, Schema Guided Dialogue (SGD) \citep{rastogi20sgd}, Action Based Conversations Dataset (ABCD) \citep{chen21abcd}, Google Simulated Conversations (GSIM) \citep{shah18gsim}, and Key-Value Retrieval for In-car Assistant (KVRET) \citep{eric17kvret}. The open domain datasets include DailyDialog (DD) \citep{li17dailydialog}, Wizard of Wikipedia (WOW) \citep{dinan19wow}, Empathetic Dialogues (ED) \citep{rashkin19empathetic}, and Machine Interaction Dialog Act Schema (MIDAS) \citep{yu21midas}.
The datasets also span a variety of data collection methodologies, such as M2M or Wizard-of-Oz, which 
has a close connection to the types of noise produced. We also consider whether the interlocutors engage in real-time vs. non-synchronous chat. 
Details of each dataset can be found in Table~\ref{tab:dst} and Appendix~\ref{sec:datasets}.


The taxonomy creation process starts by uniformly sampling 1\% of conversations from each corpus, rounding up as needed to include at least 100 dialogues per dataset.  Five expert annotators then conducted two rounds of review per conversation to tally noise counts, with a third round to break ties if needed.  The group also cross-referenced each other to merge duplicate categories and resolve disagreements.
Notably, the final taxonomy purposely excludes 
sources of noise that occur less than 0.1\% of the time. This active curation supports future denoising research by focusing attention on the most prominent sources of noise.

\begin{figure*}[ht]
  \includegraphics[width=\textwidth]{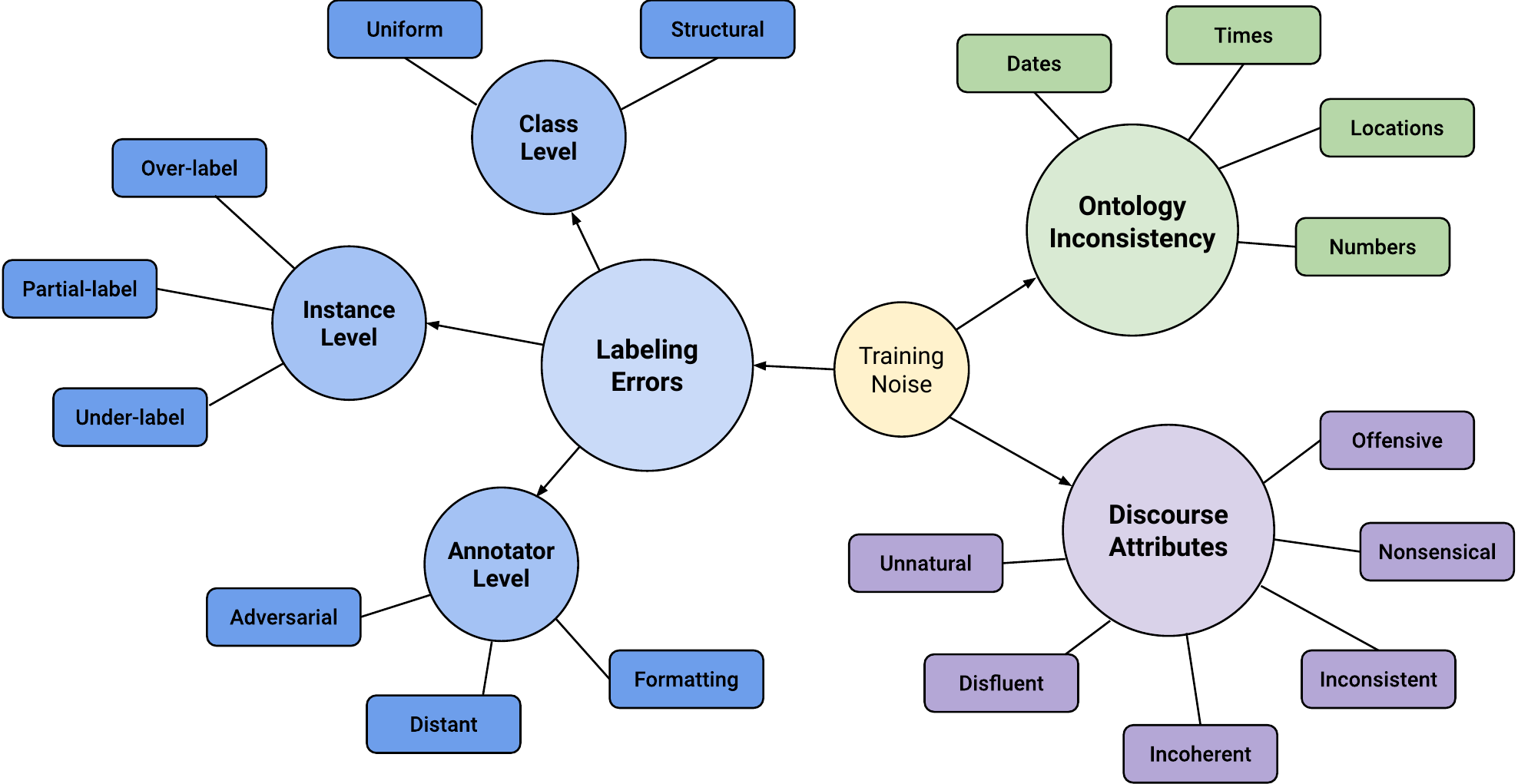}
  \caption{Diagram of the main sources of noise that affect training, based on review of the data. Our taxonomy also includes inference noise which occurs when users interact with the dialogue agent (See Fig~\ref{fig:inference}).}
  \label{fig:taxonomy}
\end{figure*}

\section{Sources of Noise}
\label{sec:taxonomy}
Through careful review of the data, we discover that dialogue systems encounter issues either from noisy training inputs during model development or from noisy user inputs during model inference.


\subsection{Training Noise}
Noisy training data impacts model learning, before any user interaction with the system. 
The sources of noise are derived from labeling errors, ontology inconsistencies or undesirable discourse attributes. 

\subsubsection{Labeling Errors}
For a given dataset of $(X,Y)$ pairs, any occasion when the target label $y$ is labeled incorrectly can be considered a labeling error. 

\paragraph{Class Level} When noise occurs due to confusion between two classes, this is considered a class-level labeling error. This can be further sub-divided into \textit{Uniform Label Swapping} or  \textit{Structured Label Swapping}.  In the former, symmetric noise implies all classes have equal likelihood to be confused with any other class, whereas in the latter certain classes are more likely to confused with other related classes. For example, ``anger'' as a label is more likely to be confused with ``frustration'' than ``joy'' when performing emotion detection.  
\paragraph{Instance Level} Noise comes from the example itself due to the complexity of interpreting natural language, which is especially common within dialogues~\citep{zhang21learning}. For example, annotators may carry over the dialogue act from the previous turn, even though it is no longer relevant, resulting in \textit{Over Labeling}.  Conversely, \textit{Under Labeling} is when a label is missed.  \textit{Partial Labels} occur when some labels are correct, while others are not.  This is common in dialogue due to the prevalence of multi-label examples, such as an utterance with two slot-values to fill.  (See Figure~\ref{fig:front_page})

\paragraph{Annotation Level} Noise arises due to the labeler or data collection process. \citep{snow04hypernym}.  Applying heuristics on a gazetteer to label named entities in NER produces \textit{Distant Supervision} noise.  Human annotators are also a source of noise either purposely from \textit{Adversarial Actors} or inadvertently from annotators acting in good faith still leading to \textit{Formatting Mistakes}.  (See Table~\ref{tab:noise_examples})

\begin{table*}[ht!]
\small
\begin{tabular}{|l|c|}
\hline & \\[-2ex]
\ \ \textbf{Dialogue}  & \textbf{Labels} \\ 
\hline & \\[-2ex]
SGD $- \ [$ Ontology Inconsistency > Date, Time $]$ & 
     \scriptsize{\texttt{GetCarsAvailable(pickup\_city=Chicago,}} \\
\qquad	User: I need a rental car in Chicago on the 3rd of this month.  &
     \scriptsize{\texttt{pickup\_date=3rd of this month)}}\\
\qquad 	System: When and for how long will you need the car? 			&  \\
\qquad  User: I'd like it from 12:30 in the afternoon till next Wednesday.	&
	\scriptsize{\texttt{GetCarsAvailable(pickup\_time=\textcolor{purple}{12:30 in}}} \\
\qquad 							$\ldots$								& 
    \scriptsize{\texttt{\textcolor{purple}{the afternoon}, dropoff\_date=Wednesday)}} \\
\qquad System: So you'd like to reserve a standard car from March 3rd at 12:30 pm &  \\
\qquad \quad until March 6th from the O'Hare International Airport location? 	& 
    \scriptsize{\texttt{ReserveCar(dropoff\_date=March 6th,}} 	      \\
\qquad User: Yes that'll work												& 
    \scriptsize{\texttt{pickup\_time=\textcolor{purple}{12:30 pm})}}\\
\hline  & \\[-2ex]
MIDAS $- \ [$ Discourse Attribute > Incoherent $]$  & \\
\qquad User: one guy                                                    & \\
\qquad Agent: what do you think about christopher nolan's acting        & Revised dialog act: \\
\qquad User: \textcolor{purple}{you can't get a boy} 										& \scriptsize{\texttt{statement} $\rightarrow$ \texttt{nonsense}} \\
\hline  & \\[-2ex]
TT $- \ [$ Labeling Error > Annotator Level > Formatting $]$ & \\
\qquad	User: We would like to see the Rhythm Section. That sounds good.& 
    \scriptsize{\texttt{(name.movie=`the Rhythm Section')}} \\
\qquad Assistant: How many tickets will you need today?                 & \\
\qquad User: We will need 4 tickets. 								    & 
    \scriptsize{\texttt{(num.tickets=4)}} \\
\qquad Assistant: Where would you like to see the movie?                & 
    \scriptsize{\texttt{(location=`San Antonio', name.theater}}\\
\qquad User: We would like to see it in San Antonio at Cinemark McCreless Market. &
        \scriptsize{\texttt{=`\textcolor{purple}{inemark McCreless Market.}'}}) \\
\hline
\end{tabular}
\caption{Selected qualitative examples of dialogue noise. Best viewed in color.  Many more examples in Appendix~\ref{sec:examples}. } \label{tab:noise_examples}
\end{table*}

\subsubsection{Ontology Inconsistency}
Another source of noise comes from inconsistent formatting when constructing the ontology. The only entities which actually contained issues are
(a) \textbf{Dates}: tomorrow, Jan 3rd, 1/3/2022, January 3 
(b) \textbf{Times}: 14:15, 2:15 PM, quarter past 2, 215pm
(c) \textbf{Locations}: NYC, New York, ny, the big apple
(d) \textbf{Numbers}: three, `wife daughter \& I', 3, `Me and my two buddies'.
In contrast, inconsistent names (ie. Fred Miyato, Mr. Miyato, fred miyato, my father) only occurred occasionally.  
Lack of standardization in the ontology was so pronounced in certain datasets that classifying labels becomes untenable, leaving generation or copying as the only viable method of predicting slot-values.

\subsubsection{Discourse Attributes}
Dialogue agents developed for response generation often mimic the behavior found in the training examples, so one hopes they contain positive discourse attributes while avoiding negative ones.  We identify six such attributes by following qualitative metrics commonly used for dialogue evaluation and through our own review of the conversations. 

(1) \textbf{Fluent} utterances flow well, obey proper grammar, and are syntactically valid.
(2) \textbf{Coherent} dialogues are semantically valid, and make sense such that they are interpretable and understandable by a general audience.  
(3) \textbf{Consistent} models do not contradict what was stated earlier in the conversation, or haphazardly change their stance on a subject.
(4) \textbf{Sensible} models follow common sense principles and understand basic natural laws (ie. gravity).
(5) \textbf{Polite} dialogue models avoid toxic language or offensive speech.  They should not exhibit overt bias towards certain groups or minorities. 
(6) \textbf{Natural} dialogues reflect how people generally talk in real life. In addition, the speakers should not break the fourth wall by directly or indirectly referring to the data collection process.  



\subsection{Inference Noise}
Inference noise refers to issues that occur in test time, during user interaction with the system after deployment to production.
This aligns nicely with the concept of out-of-scope errors~\citep{chen21gold}, which are made up of two categories: out-of-distribution cases and dialogue breakdowns.

\subsubsection{Out-of-Distribution (OOD)}
Causes of OOD \citep{peng20raddle} include:
\paragraph{Novel queries} The user asks the model to do something it was not trained to do. Example: the customer asks about frequent flyer miles, but the agent is only capable of making or modifying flight reservations.  
The model fails for these requests since it was never taught to handle such queries.
\paragraph{Unseen entities} Facing new entities or values not seen during training.  Although difficult, we could still expect a model to understand a portion of such queries by generalizing from the context. 
For example, ``I would like a flight from Miami to Puffville''. Even though the model has never heard of `Puffville', it can infer from context that this is the desired value for the destination slot. 
\paragraph{Domain shift} The dialogue system must make predictions in a new domain (taxi vs. flight).  Commonly tackled in zero-shot settings, we can expect models to occasionally generalize because there may be shared slots across domains (ie. departure time is shared by both taxi and flights queries). 

\begin{figure*}[ht]
  \includegraphics[width=\textwidth]{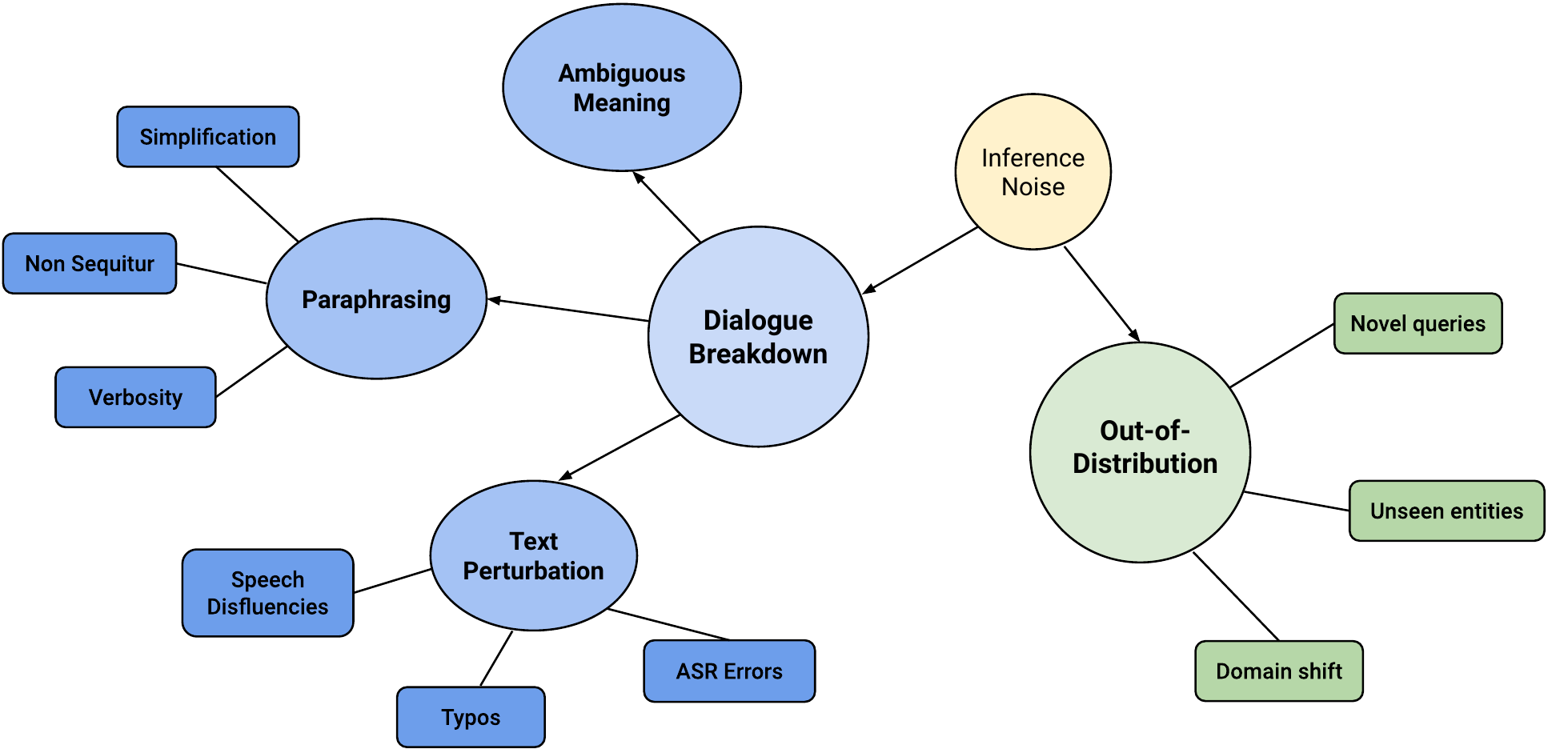}
  \caption{Diagram of the sources of noise that affect dialogue systems during inference.}
  \label{fig:inference}
\end{figure*}

\subsubsection{Dialogue Breakdown}
In contrast to OOD issues, dialogue breakdowns are situations a model should be able to handle since the scenario is within the bounds (i.e. in-domain) of what the model was trained to understand~\citep{higashinaka16breakdown}.  However, due to noise from ambiguous or unclear user input, communication breaks down and the conversation is unable to continue. \citep{higashinaka15taxonomy}.

\paragraph{Ambiguous Meaning}
Query or statement that the model should be able to handle, but caused confusion, possibly because the model failed to take the dialogue context into consideration. For example, a co-reference issue may cause difficulty in interpreting the user intent.  “Yea, let’s go with that one” is unclear when viewed in isolation.  To resolve this type of noise a model should look at the broader conversational context.

\paragraph{Paraphrasing}  The text is rephrased to become: (a) \textit{Simplification}: request may be simplified or shortened that makes it unclear what the user wants. 
(b) \textit{Non Sequitur}: response is plausibly in-distribution, but does not reasonably answer the question.
(c) \textit{Verbosity}: request is so verbose that the underlying request is lost. (See Appendix~\ref{sec:paraphrase})

\paragraph{Text Perturbations} Notable instances include (a) \textit {ASR Errors} that fail to ``wreck a nice beach'' (recognize speech) (b) \textit{Typos} and other syntax errors on the user input.  This is distinct from formatting mistakes by annotators, which are errors on the target output. (c) \textit{Speech Disfluencies} such as repeats, corrections, or adding `umm' to start a utterance \citep{liu20robustness, peng20raddle}.
\section{Noise Patterns}
Beyond categorization, manually reviewing 10K+ utterances also provides unique insights. 

\paragraph{How often does noise appear?}
The percentage of dialogues with at least one instance of noise comes out to an average of 11.2\%, a median of 10.6\%, with a standard deviation of 3.7\%.  However, given the approximate nature of sampling, the extra digits may not be significant.  Instead, we assert the rate of noise in curated dialogue datasets is usually over 5\%, rarely above 20\% and typically around 10\%.
Since these rates are relatively low, denoising techniques aiming to combat extremely high levels of noise may be impractical. 

\paragraph{What noise types are most common?}
While most existing denoising algorithms are designed to resolve class confusion~\citep{sukhbaatar14noisycnn, patrini17forward, goldberger17adaptation}, 
our analysis reveals that instance-level noise is actually much more common, showing up in nearly 10\% of cases compared to just 5\% for class-level errors. 
Class-level noise assumes a latent noise transition matrix stochastically switches labels from one class to another.  However, the prevalence of instance-level noise implies that the more likely explanation is that some examples are simply more confusing then others due to the genuinely ambiguous nature of dialogue~\citep{pavlick19inherent, nie20chaosnli} 

From an algorithmic perspective, the upshot is that developing denoising methods to target individual examples rather than class errors are likely to be most effective.  Furthermore, we discovered that noise is clustered rather than evenly distributed,
so filtering out or relabeling these particularly noisy instances should have an out-sized impact.


\paragraph{Why is X source of noise missing?} 
The expected influence of some sources of noise are greatly exaggerated.
Building out the taxonomy not only shows the most likely sources of noise, but equally notable is uncovering the \textit{least} likely noise types. 
Concretely, the threat of adversarial actors is largely overblown~\citep{dinan2019build}, as spam-like activity appears less than 2\% of the time.  Offensive speech is the subject of numerous dialogue studies~\citep{khatri2018offensive, xu2021bot, sun2022safety}, but is practically non-existent in reality (<0.5\% of cases). While hate speech may be a problem when training on raw web text~\cite{schmidt2017survey}, our empirical review reveals that
toxic language is exceedingly rare in curated datasets. Instead, unnatural utterances generated by crowdworkers role-playing as real users occurs much more often. (Full breakdown in Appendix~\ref{sec:breakdown}) 
 
Other types of noise occur so infrequently that they are missing from the taxonomy completely! Noteworthy options include inconsistent \textit{names} or \textit{titles} within the ontology (See Appendix~\ref{sec:ontology}), as well as \textit{improper reference} texts for dialogue generation tasks.  While these noise types are possible, they did not occur in practice.  We intentionally exclude all such candidates from the taxonomy since the aim is not to be comprehensive, but rather to highlight where researchers should spend their efforts.

\paragraph{Where does noise come from?} 
Our survey found that each data collection method had a propensity to produce certain kinds of noise.  This suggests noise arises as a result of how examples are annotated, rather than other factors such as conversation length (number of utterances) or dialogue style (open-domain vs. task-oriented).  For example, positive discourse attributes are most common with Post-conversation Annotation and Live Chat, which involve two human speakers engaging in real dialogue.
Wizard-of-Oz datasets are less time-consuming to produce, but contain more label noise.
In contrast, dialogues from Machine-to-Machine or Dialog Self-play (ie. starting with the labels to generate the dialogue) contain fewer label errors, but also sound less natural. 
Separately, annotator and ontology issues can be mitigated with well-written agent guidelines and proactive crowdworker screening.  
Thus, practitioners should consider these noise trade-offs when collecting dialogue data.

\begin{table*}
\centering
\resizebox{\textwidth}{!}{%
\begin{tabular}[width=\linewidth]{r|clll|c}
\toprule
\textit{Noise Source}   &  \textbf{MultiWoz}  &  \textbf{\quad \ \ Dataset 2}
                & \textbf{\quad \ \ Dataset 3} & \textbf{\qquad Dataset 4}  &  \textbf{Average}\\
\midrule
Label Noise by Class	& 84.1 (0.13\%) & 75.8 $(0.37\%)^{DD}$	& 58.1 $(1.15\%)^{ED}$	 
                        & 78.8 $(1.92\%)^{MIDAS}$	            & 0.89\% \\
Label by Instance		& 59.1 (4.88\%) & 82.4 $(3.03\%)^{SGD}$	& 72.9 $(0.96\%)^{TT}$ 	 
                        & 98.9 $(0.12\%)^{GSIM}$	            & 2.25\% \\
Label by Annotator		& 58.2 (18.1\%) & 73.6 $(3.36\%)^{DD}$ 	& 90.2 $(1.43\%)^{TT}$ 	 
                        & 44.7 $(15.9\%)^{WOW}$	                & 9.68\% \\
Discourse Attributes	& 62.9 (9.31\%) & 36.8 $(8.42\%)^{WOW}$	& 25.6 $(5.08\%)^{ABCD}$ 
                        & 39.2 $(10.7\%)^{KVRET}$	            & 8.38\% \\
Ontology Inconsistency	& 61.9 (3.41\%) & 98.7 $(0.40\%)^{GSIM}$& 58.7 $(26.8\%)^{ED}$ 	 
                        & 84.9 $(0.94\%)^{SGD}$	                & 7.89\% \\
Out-of-Distribution		& 48.1 (28.9\%) & 83.2 $(2.04\%)^{SGD}$	& 83.3 $(10.5\%)^{ABCD}$ 
                        & 74.6 $(23.6\%)^{SGD}$                 & \textbf{16.3\%} \\
Dialogue Breakdown		& 61.8 (11.3\%) & 49.8 $(4.02\%)^{WOW}$	& 4.07 $(4.44\%)^{ED}$ 	 
                        & 72.1 $(2.08\%)^{TT}$		            & 5.45\% \\
\bottomrule
\end{tabular}}
\caption{Performance across various datasets when injected with 10\% noise.  Scores in parentheses are the percent degradation when compared to the clean version of the data. Datasets 2-4 contain a superscript representing the dataset name as described in Table~\ref{tab:dst}.  Please see Appendix~\ref{tab:mapping} for the exact task and dataset mapping for each item.} \label{tab:results}
\end{table*}

\section{Experiments and Results}
This section explores to what degree various models and dialogue tasks are impacted by each of the seven different 
categories of noise outlined in Section \ref{sec:taxonomy}. To study this, a model is trained on a clean version of the dataset and on a corrupted version with either natural or injected noise.  The level of corruption for all trials is held constant at 10\% to allow for comparison across noise types.  Datasets for each noise type are selected to maximize the overall variety, while always keeping one instance of
MultiWOZ 2.3 to aid comparison. Intuitively, sources of noise that induce a larger gap in models trained on cleaned versus corrupted data are more significant, and consequently deserve more attention as targets to denoise.

\subsection{Task Setup}
All trials are conducted with GPT2-medium as a base model~\citep{brown20gpt2}. The chosen tasks are: (1) \textit{Conversation Level Classification} (CLC) -- Choose from a finite list of labels for each conversation. (2) \textit{Turn Level Classification} (TLC) -- Make a prediction for each turn that contains a label. (3) \textit{Dialogue State Tracking} (DST) -- Predict the overall dialogue state, which may contain multiple slot-values or no new slot-values at all.  Individual values come from an enumerable or open-ended ontology. (4) \textit{Response Generation} (RG) -- Produce the agent response given the dialogue context so far. 
(5) \textit{Information Retrieval} (IR) -- Find and rank the appropriate information from an external data source, such as a knowledge base (KB) or separate document.
Metrics were chosen to adhere to the evaluation procedure introduced with the original dataset or from related follow-up work.



\subsection{Noise Injection}
For each noise category, we start by independently sampling 10\% of the data, adding the corresponding noise and training a model to convergence.  For example, consider instance-level label errors applied to MultiWOZ.  This dataset contains 113,556 total utterances so 11,356 of them are selected for corruption.  Next, one of the three sub-categories of instance noise are chosen uniformly at random.  Over-labeling occurs when a label that has recently appeared in previous turns is no longer valid. To match this behavior, we keep a running tally of recent slot-labels and occasionally insert an extra one from this pool into the current training example.  Partial-labeling is achieved by replacing a slot-label with a randomly selected one from the recent pool, and under-labeling is achieved by simply dropping a slot-label from the example.  Finally, a model is trained with the noisy data applying the same hyper-parameters as the ones used for training the standard, original model. This process is repeated for each other noise type, with details for each source of noise found in Appendix~\ref{sec:injection}.

\begin{table}  
\small \centering
    \subfloat[model][Performance on MultiWOZ for each model]{
        \begin{tabular}{lccc}
        \toprule
        & \textbf{RoBERTa}      & \textbf{GPT2} & \textbf{BART} \\
        \midrule
        Original &  45.7	    & 61.9         & 62.3	        \\
        Noised   &  39.4        & 59.1         & 61.4           \\
        \bottomrule
        \end{tabular}
        \label{tab:by_model}}
    \
    \subfloat[task][Change in performance for each task due to noise.]{
        \begin{tabular}{lccccc}
        \toprule
        & \textbf{CLC} & \textbf{TLC} & \textbf{DST} & \textbf{RG} & \textbf{IR} \\
        \midrule
        Median  & 3.4\%    & 0.9\% & 4.0\% & 10.3\% & 8.4\%  \\
        Average & 6.5\%    & 4.6\% & 8.4\% & 10.5\% & 8.1\%  \\   
        \bottomrule
        \end{tabular}
        \label{tab:by_task}}
\caption{Breakdown by dialogue task and model type}
\label{tab:breakdown}
\end{table}

\subsection{Main Results}
Denoising methods targeting class-level noise may have minimal impact since it turns out such label errors are not all that damaging with just 0.89\% drop in performance.  
On the other hand, annotator noise is quite powerful causing a 9.7\% disturbance and should be mitigated whenever possible. Luckily, our manual review showed that spamming behavior occurs infrequently in reality simply by following some best practices\footnote{For example, gold checks insert questions with known labels; timers ensure adequate time is spent on each task.}.
Negative discourse attributes can also cause major harm leading to a 8.4\% gap. 

Moving onto inference noise, ontology issues are not only quite common, but also have meaningful impact on performance, causing a 7.9\% drop. Dataset creators can ameliorate this by deciding on an ontology upfront, rather than creating one after the fact. Dialogue breakdowns also cause noticeable degradation, but the impact of OOD is most prominent among all noise types.  Neural networks are powerful enough to learn from any training signal, even complete random noise~\citep{zhang17rethink}. However, OOD cases are by definition areas the network has not seen, leading to poor performance. Data augmentation and other robustness methods may serve as a strong tool to cover the unknown space by maximizing the diversity of the examples~\citep{ng20ssmba, chen22dataaug}.

\subsubsection{Task Breakdown}
In order to study tasks across noise types, we look at the percentage change between models, 
rather than absolute difference.  Furthermore, to minimize the influence of outliers, we emphasize the median of change, rather than the average.  The results in Table~\ref{tab:by_task} show that RG and IR observe the largest drops when noise is added. 
Somewhat surprisingly, CLC has larger performance shift than TLC despite being an easier task.  We hypothesize this is because CLC examples only occur once for each conversation, whereas TLC examples occur at every turn, leading to an order of magnitude less data.  Training with the existence of noisy data depends on both the rate of noisy data as well as on a minimum number of clean examples.

\subsubsection{Model Robustness}
Prior work has suggested that models behave differently when faced with distinct types of noise~\citep{belinkov18noise}. In addition to GPT2-medium (345M parameters), we also consider a masked language model in RoBERTa-Large (355M parameters)~\citep{liu19roberta} and a sequence-to-sequence model with BART-large (406M parameters)~\citep{lewis20bart}.  These are selected due to having a comparable number of training parameters. 
Based on the results in Table~\ref{tab:by_model}, RoBERTa is the weakest performer of the group.  We hypothesize this is because many dialogue tasks are generation based, whereas BERT-based models typically perform well on classification.  Conversely, BART deals quite well with noise, suggesting encoder-decoder models as reasonable starting points for future dialogue projects. 


\subsection{Amount of Noise}
\label{sec:amount}
\begin{figure}
  \includegraphics[width=\linewidth]{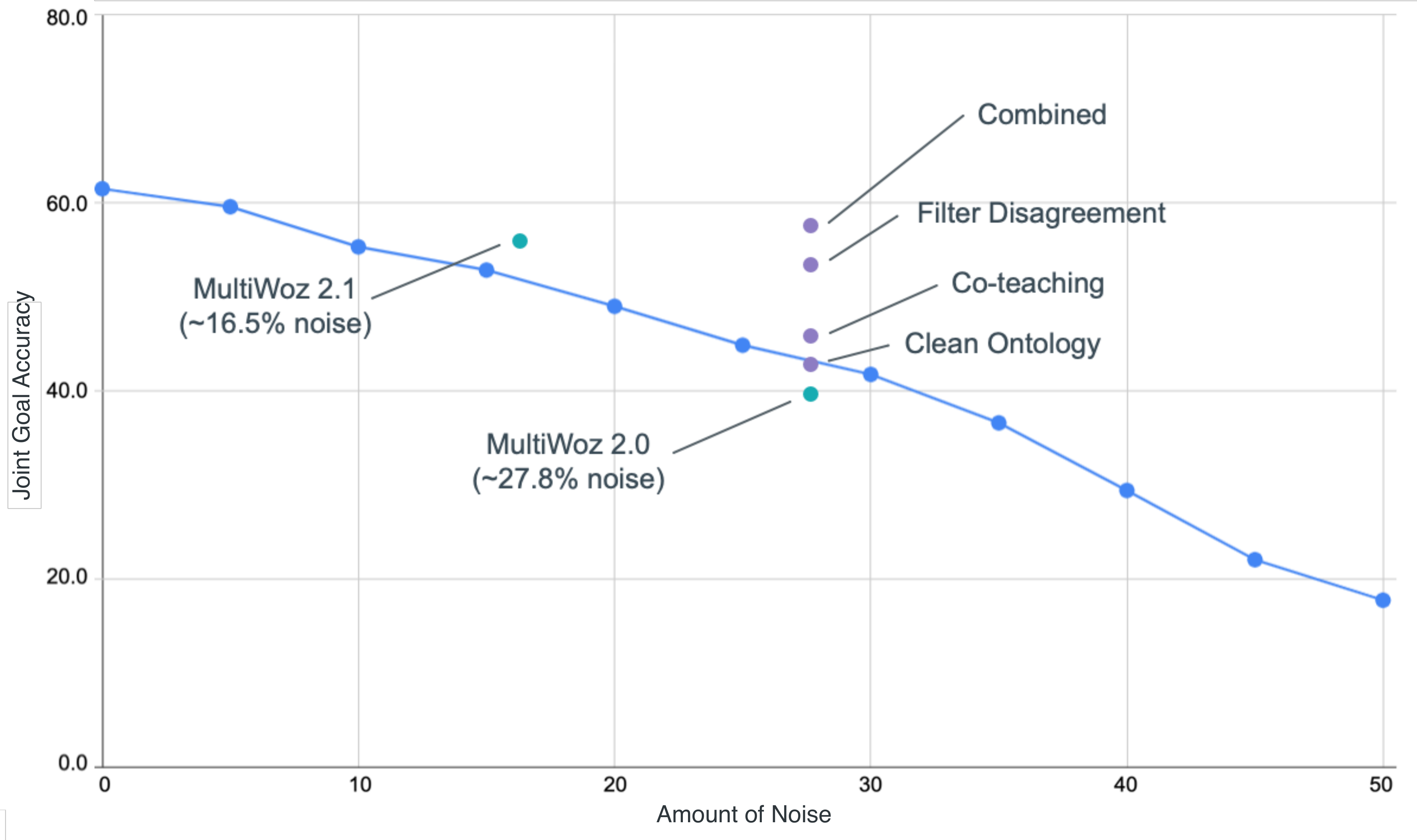}
  \caption{Impact of injecting different amounts of label and discourse noise to MultiWOZ dataset.  MultiWOZ 2.3 is defined to be 0\% noise.  MultiWOZ 2.0 and 2.1 have estimated noise levels based on the ratio of labels that are changed compared to MultiWOZ 2.3 data.}
  \label{fig:levels}
\end{figure}

We simulate increasing levels of noise by adding instance-level label errors and incoherent discourse attributes to the MultiWOZ 2.3 dataset~\citep{han21mwoz2_3}, which we define to be noise-free.  We additionally plot the performance of models trained on MultiWOZ 2.0 \citep{budzianowski18multiwoz} and 2.1~\citep{eric20mwoz2_1}, where all models are evaluated on the MultiWOZ 2.4 test set~\citep{ye21mwoz2_4}. Looking at Fig~\ref{fig:levels}, we first note that scores on naturally noisy data from MWOZ 2.0 and MWOZ 2.1 fall close to the plotted trajectory, lending credence to the overall trend.  Furthermore, we notice that as we vary the amount of noise, model performance decreases logarithmically, but surprisingly does not have a tipping point at which it fails to converge. 

\section{Dialogue Denoising}

Informed by our understanding of the sources of dialogue noise, we now design a preliminary 
denoising algorithm for learning in the presence of noisy labels. We select MultiWOZ to serve as our testbed not only because it is one of the most popular dialogue datasets, but also because it is representative of how noise affects most datasets in general (see Figure~\ref{fig:impact}).  While our method produces promising results, our aim is not to declare the noise issue solved, but rather to establish a baseline others can further improve. (More details in Appendix~\ref{sec:procedure}.)


\subsection{Algorithm}
Based on analysis in Section~\ref{sec:taxonomy}, 
MultiWOZ 2.0 is most plagued by three types of errors: ontology inconsistencies, instance label errors and out-of-distribution issues.  We now devise three solutions to resolve each source of noise accordingly.


(1) To clean up the \textit{ontology}, we drop values that do not conform to the correct format, 
and remove the associated examples from training. For example, if \texttt{time\_of\_day} slot expects the HH:MM format, then we remove all values referencing day formats (e.g. Friday).
(2) To deal with label errors, we filter out individual \textit{instances} where the predicted label from a pre-trained GPT2-medium model disagrees with the annotator label~\citep{cuendet07automatic,jiang18mentornet, chen19incv}.
We calibrate the model with temperature scaling to prevent it from being over-confident in its predictions~\citep{guo17calibrate}. 
(3) To counteract issues caused by \textit{OOD}, we augment our training data by pseudo-labeling the examples stripped out in the first two steps. When the model used for filtering is also used for pseudo-labeling, biases may propagate across each iteration.
As a result, inspired by co-teaching~\citep{han18coteaching}, we instead use a different BART-base model for pseudo-labeling to force divergence of model parameters and avoid errors from accumulating.

\subsection{Denoising Results}
We once again evaluate with MultiWOZ 2.4 since this is the cleanest version of test data. 
As seen in Figure~\ref{fig:levels}, we are able to outperform MultiWOZ 2.0 (39.8) by 16.9\% absolute accuracy and 42.5\% relative accuracy.  Ontology Clean (43.2), Filter Disagree (53.7) and Co-teaching (46.7) all show marked improvement over the original baseline, but Combined (58.6) does the best overall, reaching a score that even surpasses MultiWOZ 2.1 (56.5). These initial efforts show our ability to successfully identify and counteract sources of noise within MultiWOZ, which we encourage others to build upon.
\section{Related Works}
Our work is related to efforts to categorize noise within speech and dialog. \citet{clark96errors} proposed a theory of miscommunication consisting of channel, signal, intention and conversation where each of the four levels serves as a potential vector for noise.
Others have also studied noise in spoken dialogue systems, where they found that the main culprit stems from errors in speech transcription~\citep{paek2003toward, bohus2007error}.
Rather than a high-level framework of general communication, our hierarchical taxonomy focuses on understanding the multiple layers of noise found in written text.  

More recent works on dialogue noise discuss robustness to noisy user inputs, whereas we expand this view to also analyze noisy training inputs. \citet{peng20raddle} introduce RADDLE as a platform which covers OOD due to paraphrasing, verbosity, simplification, and unseen entities, as well as general typos and speech errors. 
\citet{liu20robustness} create a robustness benchmark which considers paraphrasing through word perturbations as well as speech disfluencies. 
Lastly, \citet{krone21robustness} considers noise from abbreviations, casing, misspellings, paraphrasing, and synonyms. 

\subsection{Survey of Denoising Methods}
Most prior works exploring learning with noisy labels were originally developed for the computer vision domain \citep{smyth94venus, mnih12aerial, sukhbaatar14noisycnn}. 
Some methods model the noise within a dataset in order to remove it, often through the use of a noise transition matrix~\citep{dawid79maximum, goldberger17adaptation}.  
Others have designed noise-insensitive training schemes by modifying the loss function \citep{rooyen15unhinged, ghosh17mae, patrini17forward},
while a final set of options manipulate noisy examples by either reweighting or relabeling them. \citep{reed14bootstrap, jiang18mentornet, li20dividemix}.
While denoising work certainly exists for NLP~\citep{snow08nlp, raykar09whomtotrust, wang19lwnl}, none of them specifically touch upon the dialogue scenario. 


\subsection{Denoising by Source of Noise}
\label{sec:recommend}
To support the effort of designing improved algorithms for combating dialogue-specific noise, we highlight potential methods that can be adapted to deal with the noise categories identified by our taxonomy in Section~\ref{sec:taxonomy}.  
To start, a common technique for dealing with \textit{class-level} errors is to learn a noise adaptation layer to recognize label noise~\citep{goldberger17adaptation}.  For \textit{instance-level} noise, besides filtering by disagreement, core-set selection~\citep{mirzasoleiman20coreset} or the Shapley algorithm~\citep{liang20Herald} can be used to identify important datapoints and thereby remove the noisy ones.  Modeling the likelihood of \textit{annotator-level} error in order to reverse its impact is also worth considering~\citep{welinder10crowds, hovy13mace, guan18whosaidwhat}.  Next, a model trained on NLI data can be used to screen out inconsistent \textit{discourse} examples~\citep{welleck19dnli}. A model trained on Prosocial Dialogue data can learn to reduce toxicity~\citep{kim22prosocial}. 
In terms of \textit{discourse fluency}, one can train a student model to reweight its logits during inference based on a large language model ~\cite{brown20gpt2} to improve the fluency of the student. Another method is to create an \textit{ontology} upfront which defines the allowed entities before data collection and enforcing this by having checks upon label submission. \textit{Out-of-Domain} issues can be handled with the use of more examples to increase the coverage and diversity of the solution space to limit OOD errors. This can be tackled by performing data augmentation on the in-domain ~\cite{feng21survey} or out-of-domain examples~\cite{chen21gold}. Lastly,  \textit{dialogue breakdown} can be mitigated by screening for annotators through minimum acceptance rates, language filters, and pre-qualifications quizzes (ie. quals). 

\section{Conclusion}
This paper categorizes the different sources of noise found in dialogue data and studies how models react to them.  We find that dialogue noise is divided into issues that occur during training and during inference. 
We also find that conversations pose unique challenges not found in other NLP corpora, such as discourse naturalness and dialogue breakdowns. 
Our study further reveals that the most common sources of noise are actually based on the ambiguity of individual instances, rather than systematic noise across classes or adversarial annotators actively harming data collection efforts.

Despite being surprisingly resilient, dialogue models nonetheless experience a notable drop in performance when exposed to high levels of noise.
To combat this, we design a proof-of-concept denoising algorithm to serve as a strong foundation for others to compare against. 
We apply this algorithm successfully to the MultiWOZ 2.0 dataset, raising the accuracy by 42.5\% over the original baseline. 
We hope our survey informs the collection of cleaner dialogue datasets and the development of advanced denoising algorithms targeting the true sources of dialogue noise.


\clearpage
\section{Limitations}
In terms of the noise taxonomy, one limitation is that we only consider natural language text within dialogue.  It could be useful to conduct a detailed breakdown of speech noise or multi-modal noise that occurs when collecting conversations grounded by images.  Furthermore, our effort to discover dialogue categories of noise was done by reviewing ten datasets.  Doubling the number of datasets reviewed or sampling more data would likely lead to more precise error rates, but is unlikely to change the results since the patterns were steady even half way through.  Our survey also does not include all theoretically possible sources of noise and instead is limited to actual sources of noise which we saw occur in the data.  We argue this type of taxonomy serves a more practical purpose.

Another limitation is our study of how the different types of noise affect the model.  In most cases, this was done by injecting artificial noise into the dataset to see how the model performance reacts.  In a few cases, we were able to obtain cleaned data to measure the impact of noise, but ideally this would be true for all the noise categories.  In this way, we would be able to measure not only how much noise harms performance, but also how much benefit denoising can help.

The main limitation of our proposed denoising method is that it has only been applied to the MultiWOZ dataset.  Although we have strong reason to believe in its generalizability to other settings, this has not been proven.  However, our goal is not to solve the dialogue denoising problem, but rather the encourage readers to consider designing their own improved algorithms.  In this sense, we believe the experiments in MultiWOZ provide a strong foundation for others to build on.
\bibliography{anthology,custom}
\bibliographystyle{acl_natbib}

\clearpage

\appendix
\newpage

\section{Dataset Descriptions}
\label{sec:datasets}

In no particular order, the datasets we study are:
\begin{enumerate}
\item DailyDialog (DD) - a collection of conversations from the web about everyday events,  curated for teaching English. \citep{li17dailydialog}
\item Wizard of Wikipedia (WoW) - a wizard reads an article on Wikipedia and then talks about it with their partner \citep{dinan19wow}
\item Google Simulated Dialogue (GSIM) - a large scale Machine-to-Machine (M2M) dataset build through paraphrasing, covers movie and restaurant domains. \citep{shah18gsim}
\item Action Based Conversations Dataset (ABCD) - customer service conversations that mimic agents in real-life where actions are taken to resolve customer issues based on instructions in agent guidelines  \citep{chen21abcd}
\item MultiWoz 2.0 (MWoz) - a multi-domain dialogue dataset. Note that we use the original version for initial analysis because it contains true noise, before any additional cleaning. \citep{budzianowski18multiwoz}
\item TicketTalk (TT) - As part of the third installment of TaskMaster, this dataset also uses the M2M style, but focuses on the single vertical of movie ticket booking. \citep{byrne19taskmaster}
\item Empathetic Dialogues (ED) - a set of dialogues that aim to teach models to be empathetic by being more attuned to what a user is feeling. \citep{rashkin19empathetic}
\item Machine Interaction Dialog Act Schema (MIDAS) - created for the Amazon Alexa challenge with Gunrock. Transcribed conversations are with actual Alexa users, and not crowdworkers. \citep{yu21midas}
\item Schema Guided Dialogue (SGD) - the most comprehensive DST dataset to date, with a heavy focus on slot-filling for API calls. Contains natural OOD splits.  \citep{rastogi20sgd}
\item Key-Value Retrieval for In-Car Assistant (KVRET) - Task oriented dataset with a knowledge base for querying items. Covers navigation, weather and scheduling domains. \citep{eric17kvret}
\end{enumerate}

\section{Label Error Details}

\paragraph{Class Level} Examples are labeled incorrectly due to confusion with another class.
\begin{itemize}
    \itemsep -3pt 
    \item \textit{Uniform Label Swapping}: symmetric noise where all classes have equal likelihood to be confused with any other class. The assumption is that noise is injected through a randomly initialized noise transition matrix.
    \item \textit{Structured Label Swapping}: asymmetric noise where certain classes are more likely to confused with other related classes. For example, a cheetah is more likely to be confused with leopard than a refrigerator when performing image recognition.  Alternatively, dogs and wolves are likely to be confused for each other much more often than with horses since those animals are similar to each other.
\end{itemize}

\paragraph{Instance Level} Noise comes from the example itself due to the complexity of interpreting natural language.  This is the realization that even when annotators act in good-faith, mistakes are still made since the instances themselves are difficult to label. Errors must be determined on a case-by-case basis. 
\begin{itemize}
    \itemsep -3pt
    \item \textit{Over Labeling}: annotator added a label, but should be removed since it is unnecessary. Example: carrying over a slot-value from the previous turn to the current dialogue state when it is not warranted. 
    \item \textit{Under Labeling}: annotator missed the label, when most people would include it. Example: failing to notice a newly mentioned criteria in the dialogue state.  This also includes cases where a better label could have been used, but the option is missing from the ontology and consequently prevents the example from being properly labeled.
    \item \textit{Partial Labeling}: part of the label is correct, but other parts are not.  For multi-intent utterances, the annotator may have captured one intent, but not the other. For slot-filling tasks, the annotator may have selected the appropriate value, but assigned it to the wrong slot.
\end{itemize}

\paragraph{Annotation Level} Noise arises due to the labeler or data collection process. \citep{snow04hypernym}
\begin{itemize}
    \itemsep -3pt 
    \item \textit{Distant Supervision}: the noise results from the fact that the label is not from a human, but rather weakly labeled from distant supervision \citep{sun17revisiting}.  For example, using a gazetteer for labeling named entities in NER.  As another example, you use the SQL results to train a semantic parser, rather than an annotated SQL query.
    \item \textit{Adversarial Actors}: meant to mimic spammers, this is characterized by repeating patterns or irrational behavior. For example, the annotator selects ``greeting'' dialogue act as the label for every single utterance regardless of the underlying text. \citep{raykar09whomtotrust, hovy13mace, khetan18mbem}  Other examples include bad actors in social media who provoke chatbots into producing unsafe content or labelers who mark every review as possessing positive sentiment without actually reading the passage.
    \item \textit{Formatting Mistakes}: Caused by non-experts making human mistakes, which are independent of the dialogue context. For example, typos or off-by-one errors, such as when the labeler failed to highlight the entire phrase during span selection. (See Table~\ref{tab:noise_examples}) 
\end{itemize}

\section{Ontology Inconsistency Details}
\label{sec:ontology}
Another source of noise comes from inconsistent formatting when constructing the ontology. More specifically, the creators of the dataset did not set a canonical format for each type of slot being tracked.  
While we can imagine many other slot-types causing issues, the types of errors which actually occurred in practice include: \begin{itemize}
    \item \textbf{Dates}: tomorrow, Jan 3rd, 1/3/2022, Monday, January 3, mon
    \item \textbf{Times}: 14:15, 2:15 PM, quarter past 2, 215pm
    \item \textbf{Locations}: NYC, New York, ny, the big apple
    \item \textbf{Numbers}: three, `wife daughter \& I', 3, `Me and my two buddies'.
\end{itemize}
Other ontology issues which we thought might occur more often, turn out to happen very rarely.  For example, naming inconsistency such as [Fred Miyato, Mr. Miyato, fred miyato, my father] did not really occur. Titles of people or places [Macdonalds, MickeyD's, McDonald's, mcdonalds] also were not present.
To minimize the amount of noise from ontology inconsistency, a recommendation is to declare the allowable slot-values upfront before data collection begins.

\section{Paraphrasing Examples}
\label{sec:paraphrase}

Paraphrasing can take on three general forms:

\begin{enumerate}
    \item \textbf{Simplification} -- the request may be simplified so much that it becomes unclear what the user wants.  For a restaurant scenario:
    
    \texttt{Agent: What part of town would you like to eat?} \\
    \texttt{User: W} \\ (as a shorthand for West side)

    \item \textbf{Non Sequitur} -- response is plausibly in-distribution, but does not reasonably answer the question.
    
    \texttt{Agent: What part of town would you like to eat?} \\
    \texttt{User: I would like Italian food.}
    
    Note that the user's response is still in distribution since it could have been a reasonable answer to ``What cuisine do you prefer?''.  However, in this instance, this type of response is very noisy because it fails to answer the agent's question.
    \item \textbf{Verbosity} -- the request contains extra words or entities, which makes it confusing as to exactly what the answer may be.
    
    \texttt{Agent: What part of town would you like to eat?} \\
    \texttt{User: I prefer food in the East, but I live in the South right now.}

    In this case, the user's response is not necessarily long, but it is verbose enough to make it unclear whether the user wants food in the east side of town or the south side of town.
\end{enumerate}
True paraphrasing noise should alter the text without altering the user's underlying intent.  If the text has changed so much that the user's intent has also shifted, then it should be considered adversarial behavior beyond the scope of typical dialogue noise.
    
    \texttt{Agent: What part of town would you like to eat?} \\
    \texttt{User: The Northern Lights are beautiful this time of year.}

The example above displays positive sentiment, but the user has completely ignored the agent's request.  This case borders on being incoherent and fails to move the dialogue forward. 

\section{Results Breakdown}
\label{sec:breakdown}
Aggregated amounts of noise by each sub-category:

\begin{table}[h]
\small \centering
        \begin{tabular}{l|ccc}
        \toprule
        & \textbf{Average}      & \textbf{Median} & \textbf{Std. Dev.} \\
        \midrule
        Class-level     &   4.9\%       &  3.8\%      &  0.7\%   \\
        Instance-level  &   9.7\%       &  6.9\%      &  5.4\%   \\
        Annotator-level &   1.8\%       &  0.7\%      &  2.1\%   \\
        \midrule
        Dates       &  3.6\%	   &  0.5\%      & 6.3\%   \\
        Times       &  1.1\%       & <0.1\%      & 2.0\%   \\
        Locations   &  1.3\%       &  0.3\%      & 2.1\%   \\
        Numbers     &  2.3\%       &  0.2\%      & 4.6\%   \\
        \midrule
        Incoherent  &  3.4\%       & 3.8\%      & 1.9\%   \\
        Disfluent   &  2.6\%       & 2.4\%      & 2.0\%   \\
        Inconsistent&  1.7\%       & 1.3\%      & 1.5\%   \\
        Nonsensical &  2.0\%       & 2.6\%      & 1.1\%   \\
        Offensive   &  0.2\%	   & <0.1\%     & 0.9\%   \\
        Unnatural   &  4.8\%       & 5.8\%      & 1.6\%   \\
        \midrule
        \textbf{Overall} & 11.2\%   & 10.6\%        & 3.7\% \\
        \bottomrule

        \end{tabular}
\caption{Breakdown across noise sub-categories}
\label{tab:noise_results}
\end{table}

\begin{figure*}
\resizebox{\textwidth}{!}{%
{\begin{tabular}[width=\linewidth]{r|cccc}
\toprule
\textit{Noise Source}   &  \textbf{MultiWoz 2.3}  &  \textbf{Dataset 2}
                & \textbf{Dataset 3} & \textbf{Dataset 4} \\
\midrule
Label Noise by Class    & MWOZ (TLC on intents)     & DD (CLC on topics)         
                        & ED (CLC on emotions)     & MIDAS (TLC on dialog acts)   \\
Label by Instance       & MWOZ (DST on slot-values) & SGD (DST w/ slot-values )  
                        & TT (DST w/ slot-values)  & GSIM (TLC on user acts)      \\
Label by Annotator      & MWOZ (RG of agent utt)    & DD (CLC on topics)         
                        & TT (TLC on APIs)         & WOW (RG on wizard utt)       \\
Discourse Attributes    & MWOZ (RG of agent utt)    & WOW (IR on wizard utt)     
                        & ABCD (IR on agent utt)   & KVRET (IR on KB entries)     \\
Ontology Inconsistency  & MWOZ (DST on slot-values) & GSIM (TLC on user acts)    
                        & ED (CLC on emotions)       & SGD (DST on slot-values)     \\
Out-of-Distribution     & MWOZ (DST on slot-values) & SGD (DST on slot-values)   
                        & ABCD (CLC on subflows)   & SGD (TLC on intents)         \\
Dialogue Breakdown      & MWOZ (RG of agent utt)    & WOW (RG on wizard utt)     
                        & ED (RG on agent utt)     & TT (DST on slot-values)      \\
\bottomrule
\end{tabular}}}
\caption{Mapping of model performance to datasets and dialogue tasks. Parentheses also includes the target of the task.  For example, `CLC on topics' means that the task is to classify the associated topic label at a conversation level, while `TLC on intents' means the task is to classify the intent of each user turn.}
\label{tab:mapping}
\end{figure*}

\section{Noise Injection Methods}
\label{sec:injection}

\paragraph{Class-level Label Errors} 
We create a noise transition matrix to mimic structured confusion.
Specifically, given a certain class label, we want to determine what is likely to be confused with it so we can substitute the current label for that other class.  To fill the noise transition matrix, we embed all class labels into bag-of-word GloVe embeddings and measure their similarity to other classes by cosine distance.  Then, for 10\% of examples, we sample an incorrect label given the original class according to the likelihood in the transition matrix. 

\paragraph{Instance-level Label Errors} To match the behavior of over-labeling, we keep a running tally of recent labels and occasionally insert an extra one from this pool into the example.  Partial-labeling is achieved by replacing a label 
from the recent pool, and under-labeling is achieved by simply dropping a random label from the example.

\paragraph{Annotator-level Label Errors} We mimic spammers who apply preset answers to every occasion without considering the actual dialogue. For the classification tasks, we assume a spammer randomly picks from one of the three most common labels for that task as the noisy target label.  For response generation tasks, we assume a spammer randomly responds with one of three generic phrases. 


\paragraph{Undesirable Discourse Attributes} We replace a subset of the utterances with noisy versions 10\% of the time.  Incoherent utterances are randomly selected sentences from other dialogues within the dataset.  Disfluent utterances are generated by shuffling the tokens within the current utterance.  Unnatural utterances are generated by selecting from a list of awkward sentences referencing the task.  

\paragraph{Ontology Inconsistency} To clean the data, we manually remove entries that do not comply to the proper format. We also merge similar categories to create more compact ontologies. Training examples that are covered by the remaining entries are considered the clean version, while the full, original dataset is considered the noisy version. 



\paragraph{Out-of-Distribution} 
Multi-domain data is divided such that training data contains a subset of domains while the test set includes examples from all domains.  Choosing the domains to exclude was straightforward for ABCD and SGD since they are given by the task design. Rather than choosing an arbitrary domain to leave out for MWOZ, we instead run the experiment once for each domain, and report the average of the five results.

\paragraph{Dialogue Breakdown} We reproduce this behavior by pre-training a paraphrase model and applying it to perturb 10\% of utterances.  Paraphrase model is trained on QQP, MRPC and PAWS corpora.

\section{Denoising Methods}
This section first dives into more detailed analysis of the noise found within MultiWOZ.  Based on this analysis, we then target these sources of noise with a novel denoising procedure.  Lastly, refer to Section~\ref{sec:recommend} where we provide recommendations of how to deal with each of the specific sources of noise introduced by the noise taxonomy.

\subsection{MultiWOZ 2.0 Noise Analysis}
\begin{figure}
  \includegraphics[width=\linewidth]{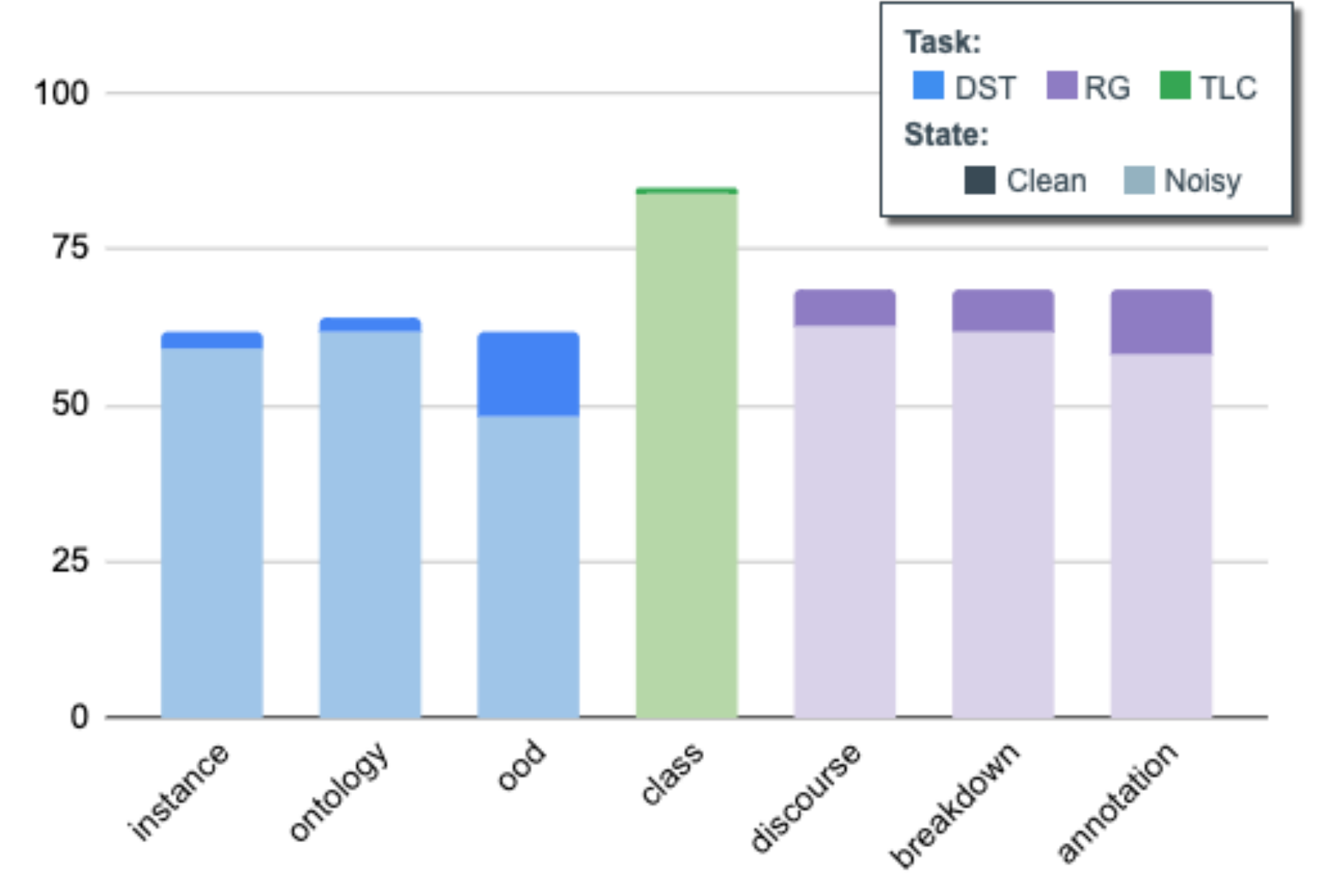}
  \caption{Impact of the different noise types on the MultiWoz2.3 dataset.  DST is dialogue state tracking, RG is response generation and TLC is turn level classification.}
  \label{fig:impact}
\end{figure}

MultiWOZ was one of the largest, most well-designed datasets upon release and still remains popular as a standard dialogue benchmark. Despite these strengths, it is also undeniable that the original dataset contained a noticeable amount of errors, which prompted the release of MultiWOZ 2.1, 2.2, 2.3 and 2.4. 
Analyzing how specific sources of noise impact MultiWOZ, Figure~\ref{fig:impact} shows that MultiWOZ is most largely impacted by OOD and Annotation-level issues.  Luckily, we found annotation-level issues to be relatively rare (~3\% of conversations) compared to instance-level (~34\%) and class-level (~1\%) labeling errors.  Each type of noise is injected into the MultiWOZ2.3 dataset at 10\% noise rate.  As usual, evaluation for all models are conducted on MultiWOZ 2.4 test set.

\subsection{Denoising Procedure for MultiWOZ}
\label{sec:procedure}
We identify the highest likelihood sources of noise for any given dataset and dealing with each one accordingly.  MultiWOZ in particular has (1) ontology issues, (2) instance level label errors and (c) out-of-distribution examples caused by low coverage in the training set.  In turn, we proceed to deal with each of these issues as follows:

(1) To clean up the \textit{ontology}, we drop values that do not conform to the correct format for their given slots, and remove the associated examples from training.  For example, if the slot is a time of day expecting the HH:MM format, then we remove all values referencing `Friday' or `afternoon' which are incorrectly formatted.

(2) To deal with possible label errors, we filter out individual \textit{instances} where the predicted label from a pre-trained GPT2-medium model disagrees with the annotator label~\citep{cuendet07automatic,jiang18mentornet, chen19incv}.

(3) Lastly, we augment our training data to counteract issues caused by \textit{OOD} cases. In order to augment, we pseudo-labeling the datapoints that have been stripped out in the first two steps.  However, the pretrained model's predictions are unlikely to be all correct, so rather than keep all the new labels, we only keep the examples where the probability of the max value crosses the 0.5 threshold.
Then, since neural networks are often over-confident, we perform calibration with temperature scaling using a $\lambda$ parameter~\citep{guo17calibrate}.  However, pseudo-labeling with the same model that is used to perform filtering causes errors to propagate which hinders performance gains.  As a result, inspired by co-teaching~\citep{han18coteaching}, we instead use a different model to force divergence of model parameters and avoid the existing biases. In more detail, we rely on a BART-base model rather than the original GPT-2 medium, which works even though BART-base has much fewer parameters.

\section{Noise as Uncertainty}
An interesting way to view the impact of noise is through the lens of Bayesian uncertainty.  In particular, aleatoric and epistemic uncertainty can be seen caused by different types of noise.  \citet{kendall17uncertainty} describe aleatoric uncertainty as uncertainty which ``captures noise inherent in the observations.''  In contrast ``epistemic uncertainty accounts for uncertainty in the model parameters which can be explained away given enough data.''

Roughly speaking, labeling errors cause epistemic uncertainty since these errors produce uncertainty in the model parameters.  If given enough clean data to train a model, the issues caused by the noisy labels should largely be erased.  In other words, epistemic uncertainty describes what the model does not know because training data was not appropriate, so by resolving the labeling errors, the training data is now appropriate and the dialogue system can be trained successfully.

On the other hand, ontology inconsistencies cause aleatoric uncertainty since they can lead to situations where it is impossible to fix the problem by altering the training data alone.  Suppose we want the dialogue model to predict the desired time for a restaurant reservation (such as 11 AM, 6PM or 8PM), but options such as `Sunday' or `afternoon' keep appearing, which are never correct.  This would make it harder for a classifier to choose the correct time.  In the degenerate case, suppose the ontology only consisted of days of the week such as `Monday', `Wednesday' or `Friday', such that the classifier would only have the ability to choose from seven incorrect options.  In this case, adding any amount of extra data (even those labeled in the correct format) would do nothing to resolve the issue since the problem itself has been modeled incorrectly.

Accordingly, a model developer should focus on eliminating certain types of noise based on the type of uncertainty they are seeing in their dialogue system.  If the model is consistently making a handful of random mistakes, then relabeling some data or collecting new data may resolve the issue.  Alternatively, if the model is a making systematic errors then looking into the ontology or data collection procedure might be a better route.

\section{Experiment Hyper-parameters}
Learning rates were tested among [1e-5, 3e-5, 1e-4, 3e-4].  Batch sizes were held constant at 72 examples per batch.  Early stopping was employed when a model failed to improve on the development set for 5 epochs in a row.  The temperature parameter for calibrating model confidence was tested in the range of $\lambda = [1.3, 1.5, 1.7, 1.9]$.  NLTK is used for calculation of BLEU score.


\onecolumn
\section{Additional Noise Examples}
\label{sec:examples}
Two examples were chosen for each of the ten datasets, giving a total of 20 examples.  Three examples can be found in Table~\ref{sec:examples} and one is found in Figure~\ref{fig:front_page}, therefore this section contains the remaining 16 examples.  The examples were carefully selected to give good coverage of the different types of noise that occurred frequently within the data. 

\begin{longtable}{|C{1.2cm}|l|p{1cm}p{7cm}|C{2.4cm}|}
\hline & & & &  \\[-2ex]
\textbf{Dataset} & \ \ \textbf{Noise Type} & & \textbf{Dialogue} & \textbf{Comments} \\ 
\hline
& & \textit{A} & Do you have any particular hobbies , Tom ?     & \\
& & \textit{B} & Oh, yes. I love playing badminton, table tennis& \\
& &            & and computer games. And playing the games      & \\
& &            & has become my favorite.                        & \textit{Original Label:} \\
& & \textit{A} & I hear many teenagers will spend hours and     & relationship \\
& &            & hours sitting at computers. And they don't     & \\
\multirow{3}{*}{DD} & Labeling Error & &
    care about their \textcolor{purple}{health} at all. How about you? & \textit{Revised label:} \\
& \textarrow \ Class level & \textit{B} & I'm no better than them. They say it's mind- & health \\
& \textarrow \ Structured & & less to sit in front of a machine all the time,    & \\
& &            & but I think it's stimulating.                  & \\
& & \textit{A} & You won't play online games full of violence,  & 
    \multirow{5}{=}{\small{Structured, rather than Uniform since many examples
    that should be `health' are systematically marked as `relationship'.}} \\
& &            &  will you?                                     & \\
& & \textit{B} & Once in a while.                               & \\
& & \textit{A} & It's no good. And I think spending several     & \\
& &            & hours each day playing computer games is a     & \\
& &            & waste of time. You might as well do something  & \\
& &            & healthy and meaningful.                        & \\
& & \textit{B} & I hope I can practice self-control again.      & \\
\hline 
& & \textit{Agent} & Hi and welcome to ACMEcorp.  How can I & \\
& &                &  help you today? & \\
& & \textit{Customer} & \ \ So I'm a premium member gold now, & 
    \multirow{7}{=}{\textit{Over Label} because the [Membership Level] action is relevant, but not actually needed for an FAQ-related flow.} \\
& Labeling Error& & wondering how that effects international & \\
& \textarrow \ Instance level  & &  shipping & \\
& \textarrow \ Over & \textit{Customer} & \ \ If it'll still be free coming from out of the country & \\
& & \textit{Agent} & Sure, I'll be happy to look into that for you. & \\
& & \textit{Action} & Searching the faq pages ... \small{\texttt{[FAQ Search]}} & \\
& & \textit{Action} & \textcolor{purple}{Membership level of ‘gold’ has been noted.} 
\small{\texttt{ [Membership Level] }} & \\
& & \textit{Action} & System action: search membership \small{\texttt{[Membership FAQ]}} & \\
ABCD & & \textit{Agent} & with gold, you'll still have to be a shipping fee, but it's reduced for international shipments. & 
    \multirow{7}{=}{\textit{Annotator} error since the mistake was likely caused by the
    agent accidentally clicking on the button.} \\
& & \textit{Customer} & \ \ Oh I see & \\
& & \textit{Action} & FAQ answer related to membership (question3) was selected.
 \small{\texttt{ [Membership 3]}} & \\
& Labeling Error & \textit{Agent} & Is there anything else I can help you with & \\
& \textarrow \ Annotator & & today?  & \\
& \textarrow \ Formatting & \textit{Customer} & \ ya was mostly wondering about how these boots coming in would be charged to me  & \\
&  & \textit{Customer} & \ \ I think you've answered my questions for & \\
& & & now thanks  & \\
& & \textit{Agent} & You're welcome, hope you have a nice day! & \\
& & & & \\
\hline
& & \textit{Apprentice} & \quad I have visited the United States. To New & \\
& & & York City, Los Angeles, and Seattle for work &  \\
& & & and vacation. Every city was unique with its & \\
& & & own culture and loved every one of them. &  \textit{Correct label}:  \\
& Labeling Error & \textit{Wizard} & I haven't been to the East coast yet, but I have &  
    \{topic: \\
WOW & \textarrow \ Instance level & & been to Los Angeles, which is Spanish for &  'Los Angeles'\}   \\
& \textarrow \ Under  & &  ``The Angels'' &   \\
& & \textit{Apprentice} & \quad Oh I never knew. The East coast always &  \\
& & & felt busier, the West coast felt more relaxed. &  \textit{Possible:} \\
& & \textit{Wizard} & Agreed! I grew up in \textcolor{purple}{Hawaii}, where the 
\textcolor{purple}{life} &   \textit{missing labels:} \\
& & & \textcolor{purple}{expectancy} is amongst the highest in the na- & \{topic:`Hawaii' \\
& & & tion. Do you like large cities or smaller towns? & `longevity'\} \\
\hline 
& & \textit{A} & Hello Mike ! Would you like a drink ? & \\
& & \textit{B} & No, thank you. I had too much to drink & \\
& & & yesterday evening. I had a \textcolor{purple}{bad hangover} this & 
     \multirow{3}{=}{\textit{Revise label}: happiness $\rightarrow$ disgust}  \\
\multirow{4}{*}{DD} & Labeling Error & & morning.  My head felt \textcolor{purple}{terrible}. (happiness) &  \\
& \textarrow \ Class level & \textit{A} & Were you celebrating something ? &  \\
& \textarrow \ Uniform & \textit{B} & Yes. It was a friend's birthday party. We drank & \\
& & & all kinds of things - beer, wine and spirits.  & \\
& & & After midnight, we even drank cocktails! & \\
& & \textit{A} & It’s a bad idea to drink a combination of & \\
& & & alcoholic drinks. You should stick with one & \\ 
& & & for the whole evening. & \\
\hline
& & \textit{User} & We would like the 7 pm show & 
    \small{\texttt{time.prefer =\textcolor{red}{7 pm}}} \\
& & \textit{Assistant} & How many tickets? & \\
& & \textit{User} & Two tickets for my husband and myself & 
    \small{\texttt{num.tickets =\textcolor{purple}{Two}}} \\
& & \textit{Assistant} & I am booking 2 tickets for Frozen 2 at the &
\small{\texttt{num.tickets}} \\
& & & Regal Santian theater at 7 PM & 
    \small{\texttt{=\textcolor{purple}{my husband and myself}}} \\
\multirow{4}{*}{TT} & Ontology& \textit{User} & Wait! My husband just texted me & \\
& Inconsistency & \textit{Assistant} & I will hold the booking & \\
& \textarrow Time, Number & \textit{User} & My husband cant get off work that early. We need to go later. & \\
& & \textit{Assistant} & What time do you want to book tickets & 
    \small{\texttt{time.prefer}} \\
& & \textit{User} & What is available after 7PM show? &
    \small{\texttt{=\textcolor{red}{after 7PM}}}\\
& & \textit{Assistant} & There is an 8:45 and 10:00 showing & \\
& & \textit{User} & We can go to the 8:45 show & \small{\texttt{time.showing =8:45}} \\
& & \textit{Assistant} & Good. I will book 2 tickets for Frozen 2 at the Regal Santiam theater at 8:45 & \\
& & \textit{User} & Thank you & \\
& & \textit{Assistant} & Can I help you with anything else? & \\
& & \textit{User} & No that is all & \\
\hline
& & \textit{A} & I've got popcorn kernels to last me through & \\
\multirow{4}{*}{ED} & Discourse & & retirement. I wonder how long they keep for.  & \\
& Attribute & \textit{B} & That is nice. & \textit{grammar} \\
& \textarrow \ Disfluent & \textit{A} & Yea, it is. Do you like popcorn? &  \textit{mistake} \\
& & \textit{B} & Yes. Why did you \textcolor{purple}{bought} that many popcorn kernels? & \\
\hline 
\end{longtable}

\begin{longtable}{|C{1.2cm}|l|p{1cm}p{7cm}|C{2.4cm}|}
\hline 
& & \textit{Agent} & Thank you for contacting acmebrands. how & \\
& & & can I help you? & \\
& & \textit{Customer} & \ \ I need to check the status of an order & \\
& & \textit{Agent} & Okay. Could I have your name or Account ID & \\
& & & please? & \\
& & \textit{Customer} & \ \ alessandro phoenix & \multirow{10}{=}{\textit{Revision:} The agent followed the `manage shipping' flow when instead
    they should have followed the `shipping status' flow, which contains \small{\texttt{[Ask Oracle]}} \normalsize  as an action} \\
& & \textit{Action} & Account has been pulled up for alessandro phoenix. \small{\texttt{[Pull-up Account]}} & \\
& & \textit{Agent} & Thank you. Do you just need to know the order's shipping status? & \\
& & \textit{Customer} & \ \ The email that I received showed it is being & \\
& Labeling Error & & shipped to the wrong address. is it being ship-& \\
ABCD & \textarrow \ Class level & & ped to  $<$street\_address$>$, jacksonville, mi & \\
& \textarrow \ Structured & \textit{Agent} & I should be able to update the order for you & \\
& & &  depending on the shipping status. do you know & \\
& & & the shipping status? & \\
& & \textit{Customer} & \ \ it just says "order received" & \\
& & \textit{Agent} & Great, i'll be able to update the address for you if you wish. & \\
& & \textit{Customer} & \ \ ok & 
    \multirow{5}{=}{These flows are often confused for each other, so the noise is structured} \\
& & \textit{Action} & Shipping status of order received has been noted. \small{\texttt{[Shipping Status]}}& \\
& & \textit{Agent} & Can you provide me with your username, email address and order id please? & \\
& & \textit{Customer} & \ \ $<$username$>$ alessandroohoenix598@email .com     $<$order\_id$>$ & \\
& & \textit{Action} & Purchase validation in progress ... \small{\texttt{[Validate Purchase]}}& \\
& & \textit{Agent} & Thank you. & \\
& & \textit{Action} & Order has been updated with change address. \small{\texttt{[Update Order]}}& \\
\hline 
& & \textit{User} & I'm looking for a special place, can you help? & \\
& & & \small{\texttt{attraction(type=\textcolor{purple}{Special})}} &
    \multirow{2}{=}{\textit{ontology}: uppercase}\\
& Ontology & \textit{System} &  I need just a little more information to help.&  \\
& Inconsistency & &   I think all places in Cambridge are special ... & \\
& \textarrow \ Location & \textit{User} & I am looking specifically for Saint John's & 
    \multirow{2}{=}{\textit{ontology}: lowercase} \\
& & & College. \quad \small{\texttt{attraction(type=\textcolor{purple}{special},}} & \\
\multirow{2}{*}{MWOZ} & & & \qquad \qquad \ \ \small{\texttt{name=\textcolor{purple}{Saint John's College.})}} &  \multirow{3}{=}{\textit{formatting}: added an extra period} \\
& & \textit{System} &  sorry i dont have that in our list. is there  & \\
& Labeling Error & & something else i can do for you? & \\
& \textarrow \ Annotator & \textit{User} & Okay, well I also need a train departing & \\
& \textarrow \ Formatting & & for Cambridge on Wednesday. & \\ 
&  & & \small{\texttt{train(dest=cambridge,day=wednesday)}} & \\
& & \textit{System} & I have several trains headed to Cambridge on  & \\
& & & Wednesday. Where will you be departing from? $<$\textit{truncated}$>$ & \\
\hline
\multirow{4}{*}{ED} & & \textit{A} & I liked the way Lebron was snatched up by &
\multirow{2}{=}{\textit{Emotion label}: impressed} \\
& Labeling Error & &  the Lakers & \\
& \textarrow \ Instance level & \textit{B} & Will be a game changer. & 
\multirow{3}{=}{\textit{Missing labels}: hopeful, anticipating} \\
& \textarrow \ Partial  & \textit{A} & \textcolor{purple}{I hope so}.  I hope they can crush the warriors &  \\
& & \textit{B} & Did the Warriors get a replacement? & \\
\hline 
\end{longtable}

\begin{longtable}{|C{1.2cm}|l|p{1cm}p{7cm}|C{2.4cm}|}
\hline
& & \textit{User} & I want to check in on March 2nd, and stay in 1 room until the 12th. & \\
& & \textit{System} & Okay, please confirm: You want to stay in New York at 1 hotel central park, checking in tomorrow, and checking out on March 12th. you also want 1 room. & 
\multirow{5}{=}{\textit{Inconsistent} since the user wants 1 room than arbitrarily changes to 3.}\\
\multirow{4}{*}{SGD} & Discourse & \textit{User} & Actually 3 rooms please. &  \\
& Attribute & \textit{System} &  Okay, confirm: You want \textcolor{purple}{3 rooms} and you check in tomorrow? & \\
& \textarrow Inconsistent, & \textit{User} &  No. I want to check out on the 13th. & 
\multirow{5}{=}{\textit{Nonsensical} since you would want to know the price of a hotel room before booking, not afterwards.}\\
& \quad Nonsensical & \textit{System} & Please confirm: You want to check out of your hotel in New York on March 13th. & \\
& & \textit{User} & Yes, works for me. & \\
& & \textit{System} & THe reservation is made. & \\
& & \textit{User} & Does it allow pets? \textcolor{purple}{How much does it cost?} & \\
& & \textit{System} & It is pet friendly, and costs \$4,488. & \\
& & \textit{User} & Thank you, i'm all set. & \\
& & \textit{System} & Have a good day. & \\
\hline 
\multirow{3}{*}{MIDAS} & Labeling Error & \textit{User} & actually &
\multirow{3}{=}{\textit{Revise label}: statement $\rightarrow$ neg answer} \\
& \textarrow \ Class level & \textit{Agent} & what was that experience like & \\
& \textarrow \ Structured  & \textit{User} & i have \textcolor{purple}{not} & \\

\hline
& & \textit{User}   & i would like a reservation for tomorrow  &
    \multirow{5}{=}{\textit{Inconsistent} because the user is not able to select any of the offered
    times despite stating that any time will do.} \\
& & \textit{System} & what restaurant , time , and number of guests?  & \\
& Discourse & \textit{User}   & sushi boat for 5 at \textcolor{purple}{any time}.  & \\
GSIM & Attribute & \textit{System} & there is three times available: 5 pm, 6.15 pm, & \\
& \textarrow \ Inconsistent  & & and 8 pm .    & \\
& & \textit{User}   & \textcolor{purple}{none of those time work}  & \\
& & \textit{System} & unfortunately i couldn't make your reservation  & \\
& & \textit{User}   & thank you , goodbye  & \\

\hline
& & \textit{User} & \textcolor{purple}{Where do I live?} & \\
& & \textit{Assistant} & Your home address is 5671 barringer street. &
\multirow{5}{=}{\textit{Nonsensical} because the user doesn't know their own home.} \\
\multirow{3}{*}{KVRET} & Discourse & \textit{User} & Please show me a quick route to come back & \\
& Attribute & & home, thank you & \\
& \textarrow \ Nonsensical & \textit{Assistant} &  In this moment there is heavy traffic but I  & \\
& & & sent on your screen the fastest route & \\
& & \textit{User} & Ok thanks. Let's go. & \\
& & \textit{Assistant} & You're welcome.  &  \\

\hline
& & \textit{Wizard} & Marriage is also called matrimony or wedlock  & \\
& & \textit{Apprentice} & \quad \multirow{1}{*}{How long ago did people start getting married?} &  
        \multirow{3}{=}{\quad \textit{Unnatural} because the wizard behaves} \\
& & \textit{Wizard} & It is a socially recognised union between spouses  & \\
& Discourse & \textit{Apprentice} & \quad 
    \multirow{1}{*}{I know that, but when did people start doing it?} & 
    \multirow{5}{=}{\quad like a bot that simply copies facts from a data- base, rather than interacting naturally.} \\
WOW & Attribute & \textit{Wizard} & From the beginning of time. Marriage varies & \\
& \textarrow \ Unnatural  & &  around the world  & \\
& & \textit{Apprentice} & \quad I don't think it has been happening since the   & \\
& &  & beginning of time. That wouldn't make sense. & \\
& &  & Why do people get married?  & \\
& & \textit{Wizard} & Really? But throughout the history of any given culture and religion  & \\
& & \textit{Apprentice} & \quad \textcolor{purple}{I'm not sure what you mean there.} & \\
\hline
\end{longtable}

\begin{longtable}{|C{1.2cm}|l|p{1cm}p{7cm}|C{2.4cm}|}
\hline
& & \textit{System} & please confirm the date is next monday at & \textit{Original label:}\\ 
& Ontology &     & 8 pm for ephesus restaurant for 3 people  & 
\small{\texttt{\textcolor{purple}{AFFIRM\_AFFIRM\_}}}\\
GSIM & Inconsistency & \textit{User}   & yes i confirm the date is next monday at 8pm & \small{\texttt{\textcolor{purple}{AFFIRM\_AFFIRM}}} \\
& \textarrow \ General &                 & for ephesus restaurant for 3 people     & \textit{Revised label:}  \\
& & & \qquad \textit{(multi-part label not needed)} & \small{\texttt{AFFIRM}} \\
\hline
& & \textit{User} & what is the weather like in the city & 
    \multirow{3}{=}{\small{\texttt{request - \{date:\textcolor{purple}{today}, loc:Alameda, weather:hail\}}}} \\
& & \textit{Assistant} & Which city do you want the weather for? & \\
\multirow{3}{*}{KVRET} & Ontology & \textit{User} & Will it hail in Alameda today? & \\
& Inconsistency & \textit{Assistant} &   Today in Alameda is gonna be drizzle, with  & 
    \multirow{3}{=}{\small{Date is annotated as “today”, but should be a day of the week such as Monday, Tuesday}} \\
& \textarrow \ Date &  & low temperature of 70F, and high of 90F & \\
& & \textit{User} & Thank you!& \\
& & \textit{Assistant} & You are welcome.  &  \\
\hline
\end{longtable}

\end{document}